\documentclass[fleqn,10pt]{wlscirep}
\usepackage[utf8]{inputenc}
\usepackage[T1]{fontenc}

\usepackage{multirow}
\usepackage{xspace}
\usepackage{amsmath}
\usepackage{amssymb}
\usepackage{cleveref}
\usepackage{mathtools}
\usepackage{subcaption}
\usepackage{enumitem}
\usepackage{makecell}
\usepackage{booktabs}
\usepackage[export]{adjustbox}
\newcommand{\stimes}{{\times}} 
\DeclarePairedDelimiter\norm{\lVert}{\rVert}%
\newcommand{\expnumber}[2]{{#1}\mathrm{e}{#2}}

\newcommand{\modelname}{\textit{U\-Net\-Flow}\xspace}
\newcommand{\bftab}{\fontseries{b}\selectfont}

\renewcommand{\vec}[1]{\text{#1}}

\newcommand{\pred}[1]{{#1}^\mathrm{p}}
\newcommand{\gt}[1]{{#1}^\mathrm{gt}}

\newcommand{\Ppred}{\pred{{P}}}
\newcommand{\Pgt}{\gt{{P}}}

\newcommand{\rev}{\color{black}}

\DeclarePairedDelimiter\abs{\lvert}{\rvert}%

\title{Abdominal Organ Segmentation via Deep Diffeomorphic Mesh Deformations}

\author[1,3*]{Fabian Bongratz}
\author[1,2]{Anne-Marie Rickmann}
\author[1,2,3]{Christian Wachinger}

\affil[1]{Technical University of Munich, Department of Radiology, Munich, 81675, Germany}
\affil[2]{Ludwig-Maximilians-University, Department of Child and Adolescent Psychiatry, Munich, 80336, Germany}
\affil[3]{Munich Center for Machine Learning, Munich, Germany}

\affil[*]{fabi.bongratz@tum.de}

\begin{abstract}
Abdominal organ segmentation from CT and MRI is an essential prerequisite for surgical planning and computer-aided navigation systems. It is challenging due to the high variability in the shape, size, and position of abdominal organs. Three-dimensional numeric representations of abdominal shapes with point-wise correspondence to a template are further important for quantitative and statistical analyses thereof. Recently, template-based surface extraction methods have shown promising advances for direct mesh reconstruction from volumetric scans. However, the generalization of these deep learning-based approaches to different organs and datasets, a crucial property for deployment in clinical environments, has not yet been assessed. We close this gap and employ template-based mesh reconstruction methods for joint liver, kidney, pancreas, and spleen segmentation. Our experiments on manually annotated CT and MRI data reveal limited generalization capabilities of previous methods to organs of different geometry and weak performance on small datasets. We alleviate these issues with a novel deep diffeomorphic mesh-deformation architecture and an improved training scheme. The resulting method, \emph{\modelname}, generalizes well to all four organs and can be easily fine-tuned on new data. Moreover, we propose a simple registration-based post-processing that aligns voxel and mesh outputs to boost segmentation accuracy.
\end{abstract}
\begin{document}

\flushbottom
\maketitle
%
%

\thispagestyle{empty}


\section*{Introduction}

\begin{figure}[b!]
    \centering
    \includegraphics[width=\textwidth]{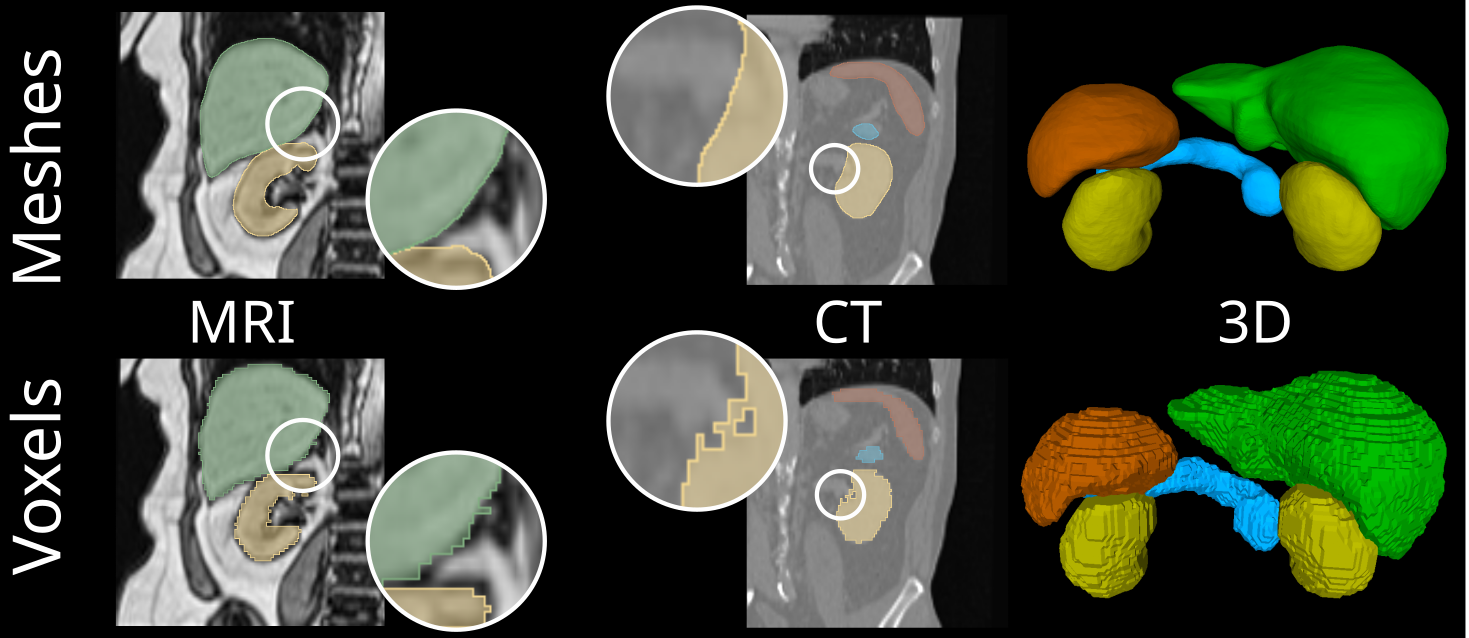}
    \caption{Meshes (top, computed with \modelname) represent shape characteristics of abdominal organs, i.e., smooth surface morphology and spherical topology, better than voxel maps (bottom), as visualized for MRI, CT, and a 3D surface rendering.}
    \label{fig:figure1}      
\end{figure}

The goal of medical image segmentation is to extract anatomical structures in the image. 
Despite the progress in tomographic imaging, the image resolution of CT and MRI is typically in the order of millimeters. 
Hence, even accurately predicted segmentation, as well as manual annotations, can only approximate the true organ contour since they are limited to the voxel grid. The consequence are stair-case artifacts shown in~\Cref{fig:figure1}. 
While voxel-based biomedical segmentation is sufficient for coarse morphological analyses, fine-grained geometric analyses of organs and clinical navigation systems need highly accurate segmentations of organ boundaries~\cite{Palomar2016planning}. Therefore, voxel-based  segmentation methods such as UNet\cite{ronneberger2015_unet} and its variants~\cite{cicek2016_3d_unet,milletari2016_vnet,isensee2020} are only of limited suitability for such applications.


With 3D meshes, we can overcome the limitations of the voxel grid by having an explicit representation of the contour, see~\Cref{fig:figure1}. 3D meshes are a natural representation of shapes as they can capture the topology and smoothness of the organ surfaces. Recent approaches for surface reconstruction from image data based on template deformations~\cite{groueix2018_atlasnet,wang2019_3dn,gupta2020_neuralmeshflow,lebrat2021corticalflow,wang2018pixel2mesh,wickramasinghe2020voxel2mesh,KONG2021deep,Bongratz_2022_CVPR} seem particularly promising for abdominal organs since anatomical priors, such as spherical topology, can be directly engraved into the template --- a property that is difficult to achieve with implicit representations\cite{mescheder2019,jeong_joon2019}. In addition, homologous points, i.e., point-to-point correspondences, of the output shapes to the template are established through the deformation~\cite{li2023spatial}. These correspondences are essential for the creation of statistical shape models and for modeling longitudinal changes.

\begin{figure}
    \centering
    \includegraphics[width=\textwidth]{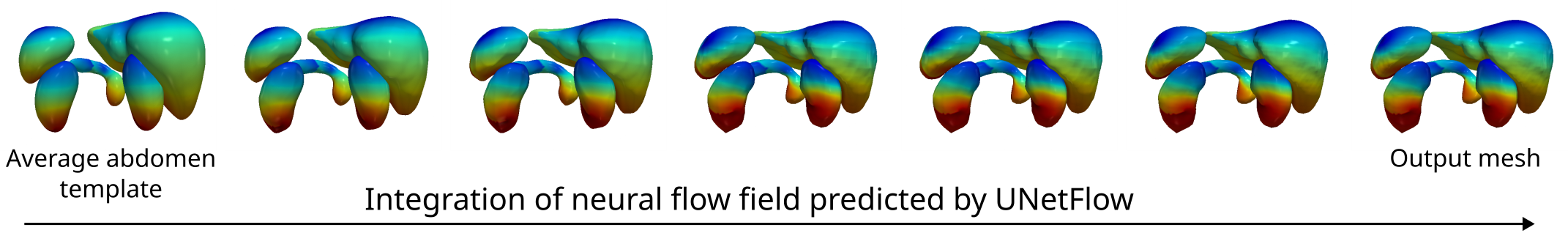}
    \caption{\modelname predicts a smooth diffeomorphic deformation of the input template, thereby establishing vertex correspondences between the template and the reconstructed shapes. Colors indicate a unique vertex ID. See the supplementary video for an animation.}
    \label{fig:flow}
\end{figure}

The origin of template-based segmentation can be traced back to the idea of active contours~\cite{kass88_snakes} in 2D and active shapes~\cite{COOTES95_activeshapes} for 3D segmentation. The main idea is to compute an ideally diffeomorphic deformation of an input template to the target surfaces as visualized in \Cref{fig:flow}. 
Recent approaches for direct surface reconstruction from image data learn these deformations from training data without requiring pre-defined landmark points. Based on the type of neural network that predicts the template warping, three groups can be distinguished. The first class of methods uses multi-layer perceptrons (MLPs) for the computation of displacement vectors~\cite{groueix2018_atlasnet,wang2019_3dn} or a neural deformation field that can be integrated numerically~\cite{gupta2020_neuralmeshflow}. The latter has the advantage that optimal integration of continuous-time deformation fields{\rev, which amounts to solving the deformation-describing ordinary differential equation (ODE),} naturally prevents self-intersections~\cite{gupta2020_neuralmeshflow}. Similarly, the second class of methods~\cite{lebrat2021corticalflow,santacruz2022_corticalflow++} relies on CNNs to predict deformation fields. A third branch of works~\cite{wang2018pixel2mesh,wickramasinghe2020voxel2mesh,KONG2021deep,Bongratz_2022_CVPR} focuses more on the graph structure of meshes. These methods combine CNNs with deep sequential graph neural networks (GNNs) that operate directly on the template mesh and can aggregate neighboring vertex features locally for computing deformation vectors. In contrast to the ODE-based approach, GNN-based deformation methods require carefully tuned regularization loss functions to enforce the smoothness of reconstructed surfaces and to avoid self-intersections. In addition, these methods predict a voxel segmentation next to the meshes. This serves as an auxiliary task for learning meaningful image features but has not yet been considered further.

In this work, we perform the first thorough evaluation of deep template-based surface reconstruction methods for joint abdominal multi-organ segmentation. This is not straightforward since most methods were developed for single shapes, and the performance typically relies on a large number of hyper-parameters. Therefore, we implement six deep mesh-deformation architectures and evaluate them on 1,000 abdominal CT scans and a smaller MRI dataset (70 scans). Our results highlight two weaknesses of existing methods: (i) corrupt pancreas segmentation due to a limited capability to generalize to organs of different geometry and (ii) difficulties training on smaller datasets. To address these issues, we develop a new deep diffeomorphic mesh-deformation method called \modelname, where we introduce a novel deep mesh supervision~(DMS) scheme and an additional voxel branch~(VB). Our experiments show that \modelname yields segmentation accuracy on par or better than existing methods while achieving the best topological measures on average over all organs. Furthermore, we analyze the relation between voxel and mesh outputs, which has been overlooked in previous works, and we showcase a simple yet effective post-processing step to improve mesh segmentation via alignment to the voxel output of the model. Finally, we introduce a fine-tuning scheme for training on small datasets and show its effectiveness on the MRI scans.

\section*{Results}

\subsection*{Homologous mesh extraction with UNetFlow}

\begin{figure}[t]
    \centering
    \includegraphics[width=\textwidth]{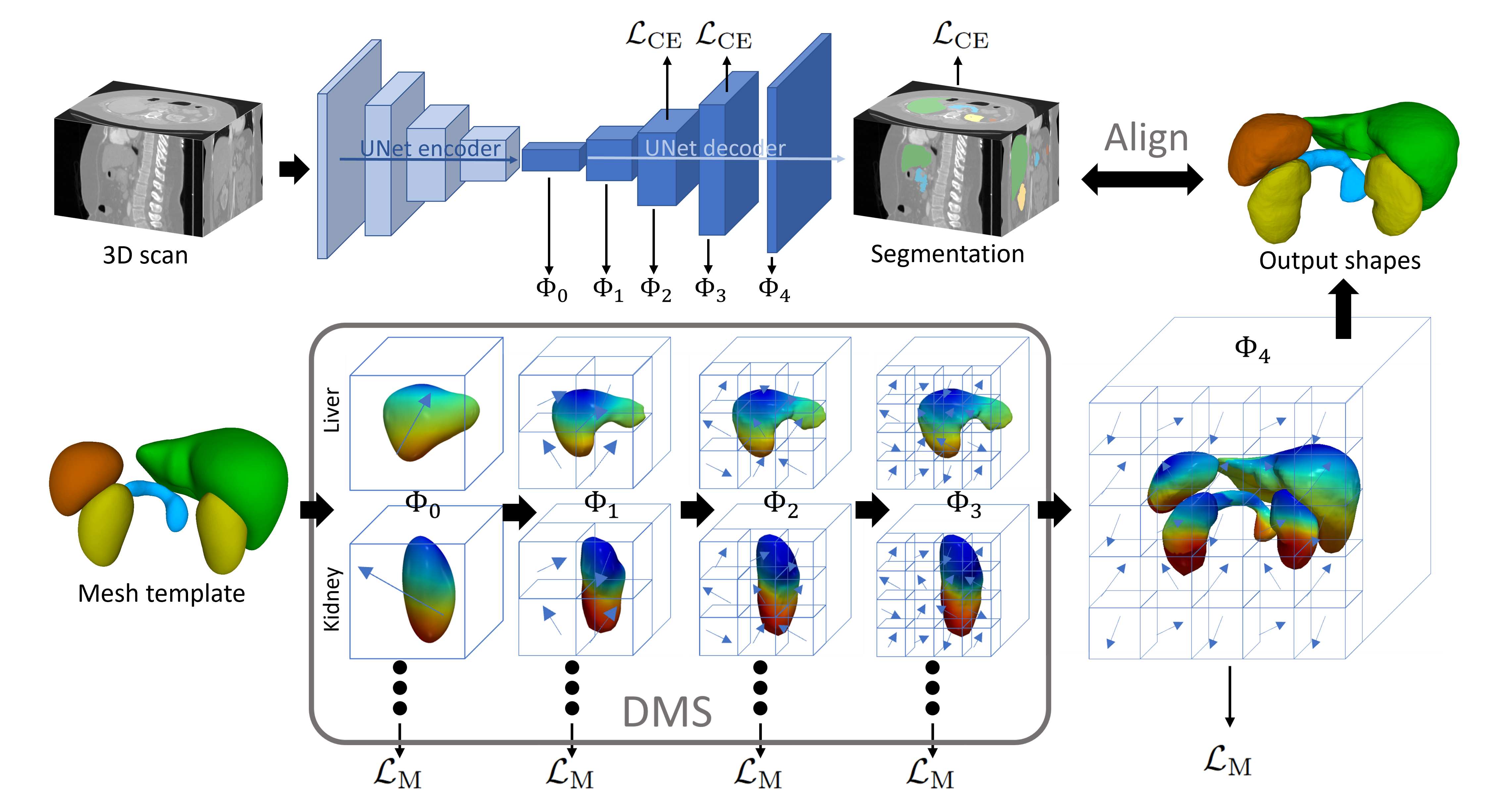}
    \caption{Architecture of \modelname. The predicted flow fields $\Phi_{0-4}$ enable a correspondence-preserving deformation of the input template, guided via deep mesh supervision (DMS).
    The model is trained with a combination of voxel ($\mathcal{L}_\text{CE}$) and mesh loss ($\mathcal{L}_\text{M}$). 
    At test time, the consistency of the output shapes with the segmentation is ensured via surface registration. Rainbow colors indicate a unique vertex ID.
    }
    \label{fig:architecture}
\end{figure}

The primary result of this work is a new deep diffeomorphic shape deformation method called \modelname. The architecture of \modelname is depicted in \Cref{fig:architecture} and further details are described in the ``Methods'' section. As input, it takes a 3D medical scan and a mesh template comprising all anatomical shapes under consideration. We pass the scan through a residual 3D UNet~\cite{isensee2020} and predict deformation fields $\Phi_{0-4}$ from the decoder. The ``resolution'' of each deformation field corresponds to the respective decoder stage, which increases the granularity of mesh deformations in a coarse-to-fine manner. To enable the joint reconstruction of multiple organs at a time, we predict an individual deformation field for each organ. An exception is given by $\Phi_4$, where we compute a single deformation field for all organs to improve the delineation of close-by surface boundaries. Formally, the warping of the input template is modeled by the {\rev ODE} $\frac{dx(t)}{dt} = \Phi(x(t), t)$, with the boundary condition at $t=0$ given by the points on the template and $\Phi$ being composed of $\Phi_0, \ldots, \Phi_4$. For the solution of this ODE, we leverage the Euler integration scheme due to its computational efficiency. Since we keep the connectivity of the meshes unchanged throughout the deformation, homologous points with an analog on the template are induced. We train \modelname with a combination of deeply supervised~\cite{zeng2017deepsupervision} voxel ($\mathcal{L}_\text{CE}$) and mesh loss ($\mathcal{L}_\text{M}$). Finally, we propose to align the output meshes to the predicted voxel segmentation for the best accuracy. 

\subsection*{Characterization of template-based mesh extraction methods}

\begin{table}[t]
 \caption{Characterization of deep mesh-deformation methods. Inference time is measured on an Nvidia A100 and includes reconstruction of all four organs except for $NMF$, which has only been implemented for the liver. The number of mesh loss terms contains the number of different loss functions computed on output meshes, ignoring terms that arise from multiple outputs.
 }
\centering
{\setlength{\tabcolsep}{.6em}
\begin{tabular}{lrccccl}

\toprule
Method & Model parameters  & End-to-end & Mesh decoder & Vertex features & Mesh loss terms & Inference time \\

\midrule


$NMF$~\cite{gupta2020_neuralmeshflow}
& 11,571,514

& $\checkmark$
& MLP 
& global
& 1
& 0.06s (liver only)
\\

$V2C\text{-MLP}$
& $6,965,772$

& $\checkmark$
& MLP
& local
& 5
& 0.52s
\\


$V2M$~\cite{wickramasinghe2020voxel2mesh} 
& $7,220,236$

& $\checkmark$
& GNN 
& local
& 4
& 0.52s
\\


$V2C$~\cite{Bongratz_2022_CVPR} 
& $7,330,828$

& $\checkmark$
& GNN 
& local
& 5
& 0.55s
\\

$CF$~\cite{lebrat2021corticalflow}
& $7,414,960$

& $\times$
& CNN 
& n/a
& 2
& 0.17s
\\



\modelname
& $6,598,736$

& $\checkmark$
& CNN 
& n/a
& 2
& 0.15s
\\

\bottomrule

 \end{tabular}
 }

\label{tab:properties}
\end{table}
\Cref{tab:properties} characterizes current deep template-based mesh extraction methods based on model parameters, end-to-end training ability, type of mesh decoder, vertex features (if any), number of loss terms, and inference time. For best comparability, we assume all methods employ the same UNet backbone and are implemented in a way to reconstruct all shapes jointly, as it is done in our implementation but usually not in the original work where the methods were proposed. 
When studying the table, we realized that $NMF$ reconstructs shapes from a compressed latent vector that represents the entire shape ``globally'' and the size of this latent representation affects reconstruction quality.
However, we are interested in comparing the different neural network architectures for mesh-based segmentation independent of the ability to learn a compressed representation of shapes. Hence, we implemented an MLP-based variant of $V2C$~\cite{Bongratz_2022_CVPR} ($V2C\text{-MLP}$), which is identical to $V2C$ besides replacing every graph layer with a linear layer. We expect that the effect of the GNN compared to an MLP can be better assessed by comparing $V2C$ to $V2C\text{-MLP}$ instead of $NMF$.


With around 6,6M parameters, \modelname has the most compact architecture. The reason is that \modelname basically consists solely of the UNet backbone, whereas the other approaches add an MLP ($NMF$, $V2C\text{-MLP}$), a GNN ($V2M$, $V2C$), or more UNets ($CF$) to it. This is also reflected by the inference time, where \modelname has a slight advantage over all other methods for reconstructing five individual shapes (liver, two kidneys, spleen, and pancreas). All methods, besides $CF$, which relies on a sequential training procedure of the UNets, are trainable in an end-to-end fashion. End-to-end training is important for transferring trained models to new data via fine-tuning and it typically makes the training less error-prone. Another characteristic is the number of mesh loss terms used for training the different architectures, ranging from one in $NMF$ to five in $V2C$/$V2C\text{-MLP}$. Note that we only consider the number of \emph{different} loss functions, i.e., geometric or regularization terms, not taking into account loss terms from different deformation stages or deep supervision. In general, it can be assumed that a larger number of loss functions makes it harder to transfer the method to new data since the weighting among loss terms is a sensitive parameter that needs to be tuned thoroughly.


\subsection*{Reconstruction accuracy and topological correctness}

\begin{table}[t]
 \caption{Comparison of deep mesh-based surface extraction methods on the CT test set ($n=147$). We report Dice scores, average symmetric surface distance (ASSD), and the relative number of self-intersecting faces (SIF). Values are (mean \textpm SD) and the best scores are \textbf{highlighted} for each organ. 
 }
\centering
{\setlength{\tabcolsep}{1em}
\begin{tabular}{lcccccc}

\toprule
&
\multicolumn{3}{c}{Kidneys} & 
\multicolumn{3}{c}{Liver} \\

Method & Dice \textuparrow & ASSD (mm) \textdownarrow & SIF (\%) \textdownarrow & Dice \textuparrow & ASSD (mm) \textdownarrow & SIF (\%) \textdownarrow  \\

\midrule


$NMF$~\cite{gupta2020_neuralmeshflow}
&  &  &  
& 0.84 \textpm 0.11 & 5.44 \textpm 4.17 & \bftab 0.00 \textpm 0.00  
\\


$V2M$~\cite{wickramasinghe2020voxel2mesh} 
& 0.89 \textpm 0.08 & 1.66 \textpm 2.19 & \bftab 0.00 \textpm 0.00
& 0.94 \textpm 0.05 & 1.93 \textpm 1.74 & 1.00 \textpm 1.38 
\\


$V2C$~\cite{Bongratz_2022_CVPR} 
& \bftab 0.91 \textpm 0.08 & \bftab 1.46 \textpm 2.33 & \bftab 0.00 \textpm 0.00
& \bftab 0.95 \textpm 0.04 & \bftab 1.51 \textpm 1.39 & 0.04 \textpm 0.25
\\

$V2C\text{-MLP}$
& 0.90 \textpm 0.08 & 1.52 \textpm 2.23 & \bftab 0.00 \textpm 0.04
& \bftab 0.95 \textpm 0.04 &  1.63 \textpm 1.63 & 0.02 \textpm 0.17 
\\

$CF$~\cite{lebrat2021corticalflow}
& 0.88 \textpm 0.14 & 2.34 \textpm 4.68 & 0.26 \textpm 1.53
& \bftab 0.95 \textpm 0.05 & 1.86 \textpm 1.75 & 0.35 \textpm 1.19

\\




\modelname
& 0.90 \textpm 0.09 & 1.74 \textpm 2.53 & 0.03 \textpm 0.42
& \bftab 0.95 \textpm 0.03 & 1.72 \textpm 0.96 & \bftab 0.00 \textpm 0.04 
\\



\midrule
& \multicolumn{3}{c}{Pancreas} & \multicolumn{3}{c}{Spleen} 
\\
\midrule



$V2M$~\cite{wickramasinghe2020voxel2mesh} 
& 0.61 \textpm 0.13 & 3.75 \textpm 3.36 & 14.15 \textpm 1.03
& 0.90 \textpm 0.10 & 1.83 \textpm 2.76 & 0.03 \textpm 0.23 
\\

$V2C$~\cite{Bongratz_2022_CVPR} 
& 0.63 \textpm 0.13 & 3.60 \textpm 3.90 & 3.88 \textpm 1.03
& \bftab 0.92 \textpm 0.09 & \bftab 1.55 \textpm 3.62 & 0.02 \textpm 0.12 
\\

$V2C\text{-MLP}$
& 0.59 \textpm 0.14 & 3.88 \textpm 3.75 & 2.42 \textpm 1.08
& 0.91 \textpm 0.08 & 1.53 \textpm 2.15 & 0.02 \textpm 0.18 
\\

$CF$~\cite{lebrat2021corticalflow}
& 0.69 \textpm 0.20 & 4.10 \textpm 4.55 & 0.26 \textpm 0.79
& 0.91 \textpm 0.13 & 2.38 \textpm 6.87 & 0.73 \textpm 2.73
\\




\modelname
& \bftab 0.72 \textpm 0.15 & \bftab 3.44 \textpm 3.51 & \bftab 0.06 \textpm 0.08
& \bftab 0.92 \textpm 0.09 & 1.73 \textpm 2.69 & \bftab 0.00 \textpm 0.03 
\\



\bottomrule

 \end{tabular}
 }

\label{tab:abdomenct1kcomparison}
\end{table}  
\begin{figure}
    \centering
    \includegraphics[width=0.75\textwidth]{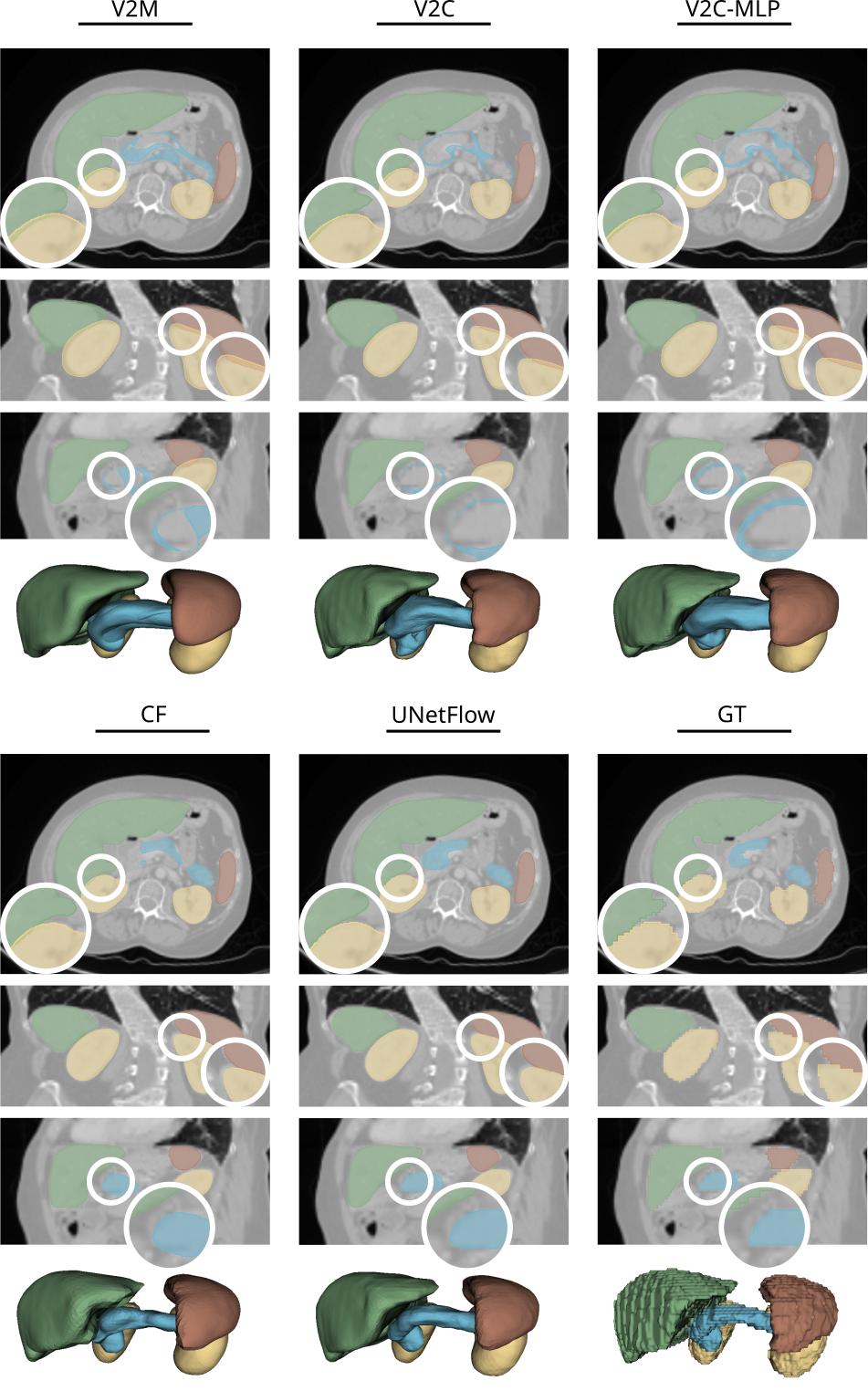}
    \caption{\modelname avoids pitfalls of previous template-based organ extraction methods, such as inter-mesh intersections and defective (``wrapped around'') pancreas shapes. Shown are axial, coronal, and sagittal views of segmentations (from top to bottom) on a scan from the CT test set and corresponding 3D views of shapes (last row, respectively).}
    \label{fig:segmentation_main}
\end{figure}
We report segmentation Dice scores and average symmetric surface distance (ASSD) on the CT test set in \Cref{tab:abdomenct1kcomparison} for all methods from \Cref{tab:properties}. We also compute the number of self-intersecting faces (SIF) per organ as a topological measure of surface quality. We observe that $NMF$ is not competitive in liver segmentation with the other methods, which we attribute to the reconstruction from a global feature vector that is probably unable to cover a high level of surface details. For kidneys, liver, and spleen, the performance of the other methods is largely comparable. For the pancreas, however, $V2C$, $V2M$, and $V2C\text{-MLP}$ yield considerably lower Dice scores. This quantitative observation is underpinned by \Cref{fig:segmentation_main}, which shows that the pancreas mesh is ``wrapped around'' multiple times in these methods. The defect is likely related to the many loss functions in these methods as it happens for the GNN and the MLP decoder similarly. \modelname and $CF$ turn out to be more robust in this regard as they achieve reasonable segmentation of the pancreas, which we attribute to the fewer loss functions and a numerical integration scheme. Notably, we find a slight but consistent advantage of \modelname (Dice of $0.87$, ASSD $2.16$mm, SIF $0.02\%$ on average over all organs) compared to $CF$ (Dice of $0.86$, ASSD $2.67$mm, SIF $0.4\%$).

\subsection*{Ablation study}

\begin{table}[htb]
\caption{\modelname ablation study on the CT validation set ($n=167$). We evaluate the impact of the proposed deep mesh supervision (DMS) and voxel branch (VB) in terms of Dice score, average symmetric surface distance (ASSD) and self-intersecting faces (SIF). Values are mean \textpm std over all organs in the validation set. 
 }
\centering

{\setlength{\tabcolsep}{.6em}
\begin{tabular}{lccccc}

\toprule
DMS & VB & Dice \textuparrow & ASSD (mm) \textdownarrow & SIF (\%) \textdownarrow \\


\midrule



- & - & 0.85 \textpm 0.14 & 2.30 \textpm 1.98 & 0.17 \textpm 0.88\\

- & \checkmark &  0.88 \textpm 0.11 & 1.92 \textpm 1.43 & 0.30 \textpm 1.31 \\

\checkmark & - & 0.88 \textpm 0.12 & 1.91 \textpm 1.29 & 0.01 \textpm 0.03 \\  


\checkmark & \checkmark &  0.88 \textpm 0.11 & 1.83 \textpm 1.29 &  0.01 \textpm 0.04
 \\

\bottomrule

 \end{tabular}
 }
 
\label{tab:ablation}
\end{table}
\Cref{tab:ablation} assesses the impact of novel architectural design choices in \modelname on the CT validation set. It can be observed that both, the proposed deep mesh supervision (DMS) as well as the added voxel branch (VB), improve the mean segmentation accuracy in terms of Dice score and average symmetric surface distance (ASSD) in average over all organs. Importantly, the DMS reduces the number of self-intersecting faces (SIF), a measure of topological correctness, by up to $97\%$ from $0.30\%$ to $0.01\%$.

\subsection*{Alignment of surface meshes and voxel segmentation}

\begin{table}[t]
 \caption{{\rev Surface} accuracy increases by rigid Iterative Closest Point (ICP) and non-rigid ICP (NRICP) registration of mesh to voxel output. We report mean values \textpm SD of average symmetric surface distance (ASSD), 99-percentile Hausdorff distance (HD99), and self-intersecting faces (SIF) on the CT test set ($n=147$). {\rev$^*$This model uses the UNet encoder and decoder from \modelname.} 
 }
\centering
{\setlength{\tabcolsep}{.5em}
\begin{tabular}{lccccccl}

\toprule
&&
\multicolumn{3}{c}{Kidneys} & 
\multicolumn{3}{c}{Liver} \\

Registration & Method & ASSD (mm) \textdownarrow & HD99 (mm) \textdownarrow & SIF (\%) \textdownarrow & ASSD (mm) \textdownarrow & HD99 (mm) \textdownarrow  & SIF (\%) \textdownarrow  \\
\midrule

- & $V2C$~\cite{Bongratz_2022_CVPR} 
& 1.46 \textpm 2.33 & 7.33 \textpm 10.5 & 0.00 \textpm 0.00
& 1.51 \textpm 1.39 & 13.1 \textpm 14.2 & 0.04 \textpm 0.25
\\

- & \modelname
& 1.74 \textpm 2.53 & 7.94 \textpm 9.76 & 0.03 \textpm 0.42
& 1.72 \textpm 0.96 & 14.0 \textpm 12.3 & 0.00 \textpm 0.04
\vspace{2pt} \\


ICP~\cite{besl1992icp} & $V2C$~\cite{Bongratz_2022_CVPR}
& 1.25 \textpm 2.20 & 6.63 \textpm 10.6 & 0.00 \textpm 0.00
& 1.50 \textpm 1.57 & 13.0 \textpm 14.5 & 0.04 \textpm 0.25
\\

ICP~\cite{besl1992icp} & \modelname
& 1.34 \textpm 2.48 & 7.07 \textpm 10.0 & 0.03 \textpm 0.42
& 1.52 \textpm 0.98 & 13.8 \textpm 12.3 & 0.00 \textpm 0.04
\vspace{2pt} \\

\rev NRICP~\cite{Amberg2007OptimalSN} & 
\rev $UNet^*$ &
\rev 4.02 \textpm 4.96 & 
\rev 22.0 \textpm 22.2 & 
\rev 0.91 \textpm 5.04 & 
\rev 2.98 \textpm 2.62 & 
\rev 31.4 \textpm 21.3 & 
\rev 0.00 \textpm 0.04 \\

NRICP~\cite{Amberg2007OptimalSN} & $V2C$~\cite{Bongratz_2022_CVPR}
& 1.16 \textpm 2.29 & 6.40 \textpm 10.3 & 0.01 \textpm 0.17
& 1.27 \textpm 0.93 & 12.6 \textpm 13.9 & 0.11 \textpm 0.60
\\

NRICP~\cite{Amberg2007OptimalSN} & \modelname
& 1.22 \textpm 2.45 & 6.62 \textpm 9.95 & 0.23 \textpm 2.71
& 1.35 \textpm 0.93 & 13.6 \textpm 13.1 & 0.05 \textpm 0.44
\\

\midrule
&& \multicolumn{3}{c}{Pancreas} & \multicolumn{3}{c}{Spleen} 
\\
\midrule

- & $V2C$~\cite{Bongratz_2022_CVPR}
& 3.60 \textpm 3.90 & 20.6 \textpm 22.8 & 3.88 \textpm 1.03
& 1.55 \textpm 3.62 & 6.79 \textpm 12.5 & 0.02 \textpm 0.12 
\\

- & \modelname
& 3.44 \textpm 3.51 & 20.8 \textpm 22.9 & 0.06 \textpm 0.08
& 1.73 \textpm 2.69 & 8.24 \textpm 8.21 & 0.00 \textpm 0.03
\vspace{2pt} \\

ICP~\cite{besl1992icp} & $V2C$~\cite{Bongratz_2022_CVPR}
& 3.68 \textpm 6.01 & 21.3 \textpm 24.9 & 3.88 \textpm 1.03
& 1.21 \textpm 1.86 & 5.54 \textpm 8.29 & 0.02 \textpm 0.12
\\

ICP~\cite{besl1992icp} & \modelname
& 3.32 \textpm 3.21 & 20.4 \textpm 23.0 & 0.06 \textpm 0.08
& 1.40 \textpm 1.91 & 7.14 \textpm 7.95 & 0.00 \textpm 0.03
\vspace{2pt} \\

\rev NRICP~\cite{Amberg2007OptimalSN} & 
\rev $UNet^*$ &
\rev 5.37 \textpm 4.01 &
\rev 34.0 \textpm 24.9 & 
\rev 1.35 \textpm 3.86 &
\rev 2.90 \textpm 3.63 & 
\rev 22.0 \textpm 18.7 & 
\rev 0.95 \textpm 5.13 \\

NRICP~\cite{Amberg2007OptimalSN} & $V2C$~\cite{Bongratz_2022_CVPR}
& 3.26 \textpm 6.73 & 20.2 \textpm 28.6 & 31.0 \textpm 4.83
& 1.00 \textpm 1.14 & 5.03 \textpm 6.98 & 0.10 \textpm 0.85
\\

NRICP~\cite{Amberg2007OptimalSN} & \modelname
& 3.12 \textpm 3.60 & 20.9 \textpm 23.8 & 0.80 \textpm 3.64
& 1.21 \textpm 2.22 & 6.25 \textpm 8.23 & 0.18 \textpm 1.39
\\

\bottomrule

 \end{tabular}
 }

\label{tab:registration}
\end{table}
$V2M$, $V2C$, and \modelname predict a voxel segmentation next to the surface meshes. However, the relation between these outputs has not been considered so far. We close this gap and demonstrate in \Cref{tab:registration} that, by using rigid Iterative Closest Point (ICP)~\cite{besl1992icp} or non-rigid ICP (NRICP)~\cite{Amberg2007OptimalSN} registration of the meshes to the voxel prediction, the segmentation accuracy of the meshes predicted by $V2C$ and \modelname can easily be improved. 
{\rev 
At the same time, the point-wise correspondence to the template is preserved by the registration. As an additional baseline, we also evaluate direct non-rigid registration of the template to the voxel segmentation produced by the UNet backbone in \modelname (a rigid registration could not cover individual shape details).

In \Cref{tab:registration}, we find that the non-rigid registration of the template to the UNet output is not competitive with the learned deformations of $V2C$ and \modelname.
For $V2C$ and \modelname, Dice and ASSD scores are generally better after the registration except for $V2C$ on the pancreas, where the ASSD is slightly higher after the application of ICP. While the improvement in accuracy is mostly larger for non-rigid registration, it also introduces more topological errors. 
On the other hand, the surface topology and, hence, the number of self-intersecting faces (SIF), remains unchanged when applying the ICP algorithm.
}

\subsection*{Generalization to MRI}

\begin{table}[t]
 \caption{Evaluation on the MRI test set ($n=18$) {\rev with models trained from scratch and, if possible, pre-trained on CT data}. We report Dice, average symmetric surface distance (ASSD), and the relative number of self-intersecting faces (SIF). Values are (mean \textpm SD) and the best scores are \textbf{highlighted} for each organ. 
 }
\centering
{\setlength{\tabcolsep}{.7em}
\begin{tabular}{lccccccc}

\toprule
&&
\multicolumn{3}{c}{Kidneys} & 
\multicolumn{3}{c}{Liver} \\

Method & \makecell{Pre-tr.} & Dice \textuparrow & ASSD (mm) \textdownarrow & SIF (\%) \textdownarrow & Dice \textuparrow & ASSD (mm) \textdownarrow & SIF (\%) \textdownarrow  \\

\midrule
$CF$~\cite{lebrat2021corticalflow} & $\times$
& 0.77 \textpm 0.09 & 3.61 \textpm 1.29 & \bftab 0.00 \textpm 0.03
& 0.89 \textpm 0.11 & 4.19 \textpm 4.19 & \bftab 0.00 \textpm 0.01 \\



\rev $V2C$~\cite{Bongratz_2022_CVPR} & 
\rev $\times$ &
\rev 0.81 \textpm 0.07 & 
\rev 3.10 \textpm 1.34 & 
\rev \bftab 0.00 \textpm 0.00 &
\rev 0.92 \textpm 0.04 & 
\rev 2.88 \textpm 1.21 & 
\rev 0.06 \textpm 0.27 \\

\rev \modelname &
\rev $\times$ &
\rev 0.80 \textpm 0.06 & 
\rev 3.21 \textpm 0.91 & 
\rev \bftab 0.00 \textpm 0.00 & 
\rev 0.92 \textpm 0.04 & 
\rev 2.73 \textpm 0.85 & 
\rev \bftab 0.00 \textpm 0.01 \\

$V2C$~\cite{Bongratz_2022_CVPR} & $\checkmark$
& \bftab 0.87 \textpm 0.07 & \bftab 1.97 \textpm 0.95 & \bftab 0.00 \textpm 0.00 
& 0.93 \textpm 0.03 &  \bftab 2.15 \textpm 0.70 & 0.08 \textpm 0.31
\\


\modelname 
& $\checkmark$
& 0.85 \textpm 0.05 & 2.45 \textpm 0.59 & \bftab 0.00 \textpm 0.00
& \bftab 0.94 \textpm 0.02 & 2.21 \textpm 0.64 & 0.06 \textpm 0.24
\\

\midrule
&& \multicolumn{3}{c}{Pancreas} & \multicolumn{3}{c}{Spleen} \\
\midrule

$CF$~\cite{lebrat2021corticalflow} & $\times$
& 0.42 \textpm 0.17 & 6.20 \textpm 2.30 & \bftab  0.03 \textpm 0.03
& 0.75 \textpm 0.12 & 4.65 \textpm 2.27 & 0.02 \textpm 0.05 \\


\rev $V2C$~\cite{Bongratz_2022_CVPR} & 
\rev $\times$ &
\rev 0.46 \textpm 0.12 & 
\rev 5.34 \textpm 2.14 & 
\rev 6.72 \textpm 1.34 &
\rev 0.80 \textpm 0.11 & 
\rev 3.59 \textpm 2.32 & 
\rev \bftab 0.00 \textpm 0.00 \\

\rev \modelname &
\rev $\times$ &
\rev 0.48 \textpm 0.16 & 
\rev 5.47 \textpm 2.20 & 
\rev 0.04 \textpm 0.04 &
\rev 0.80 \textpm 0.11 & 
\rev 3.59 \textpm 1.91 & 
\rev 0.02 \textpm 0.08 \\

$V2C$~\cite{Bongratz_2022_CVPR} & $\checkmark$
& 0.54 \textpm 0.13 & \bftab 4.10 \textpm 1.44 & 3.92 \textpm 0.87 
& 0.86 \textpm 0.08 &  \bftab 2.30 \textpm 1.31 & \bftab 0.00 \textpm 0.00
\\


\modelname & $\checkmark$
&  \bftab 0.59 \textpm 0.18 & 4.56 \textpm 2.43 &  0.05 \textpm 0.08
&  \bftab 0.87 \textpm 0.06 & 2.52 \textpm 1.17 & \bftab 0.00 \textpm 0.00
\\

\bottomrule

 \end{tabular}
 }

\label{tab:mrifinetuning}
\end{table}
Abdominal MRI is an integral part of large population studies~\cite{Littlejohns2020ukb,nako2014,Bamberg2016SubclinicalDB}. Nonetheless, the number of annotated training samples is typically smaller than in publicly available CT datasets. Therefore, we assess the generalization of template-based segmentation models to MRI. Since $CF$ relies on an intricate sequential training procedure and it is therefore unclear how to transfer a trained model to new data, we train $CF$ from scratch on the MRI data. As seen from \Cref{tab:mrifinetuning}, the results for $CF$ are far from satisfying --- with an average Dice score of 0.71 over all organs.  This is underwhelming since, although the number of training samples (47) is relatively low, this is not an atypical size for segmentation datasets and is often sufficient for segmentation models~\cite{milletari2016_vnet}. We further evaluate the best two models from \Cref{tab:abdomenct1kcomparison}, $V2C$ and \modelname, {\rev in two different settings: training from scratch on the MRI data and fine-tuning the models pre-trained on CT. It turns out that the pre-training on CT drastically improves the accuracy on MRI compared to the models trained from scratch.} The scores for the pre-trained models are mostly on par, with an advantage of \modelname in segmentation Dice on the liver, pancreas, and spleen. \modelname also shows a low number of self-intersections ($\leq 0.06\%$ for all organs) compared to $V2C$ ($3.92\%$ on the pancreas).






 

\section*{Discussion}


The automated segmentation of abdominal shapes is challenging because of
the high variability in size, shape, and position. Furthermore, point correspondences
across surfaces have to be established for vertex-wise
statistical analyses. Recent advances in template-based surface reconstruction, comprising methods like $NM$F~\cite{gupta2020_neuralmeshflow}, $V2M$~\cite{wickramasinghe2020voxel2mesh}, $CF$~\cite{lebrat2021corticalflow}, and $V2C$~\cite{Bongratz_2022_CVPR} seem to be promising for this task as they do not require any mesh extraction nor any other post-processing but directly produce meshes with point-wise correspondence to a template. However, the generalization to different shapes and imaging modalities is crucial for deploying deep learning models to clinical practice and has not been evaluated thoroughly. 

Our experiments on four organs (liver, kidneys, spleen, pancreas) and two imaging modalities (CT, MRI) uncovered the fragility of existing mesh template-based shape reconstruction methods. More precisely, methods that do not exhibit a numerical integration scheme but instead rely on regularization terms ($V2M$~\cite{wickramasinghe2020voxel2mesh}, $V2C$~\cite{Bongratz_2022_CVPR}) were unable to successfully reconstruct organs of different shape and size, which becomes evident from defective, practically useless pancreas meshes in \Cref{fig:segmentation_main}. Further, even though yielding topologically flawless liver shapes, we found $NMF$~\cite{gupta2020_neuralmeshflow} not to be competitive regarding segmentation accuracy. We suppose this is due to the global latent vector from which the organ is recovered, which is different from the localized view of the other evaluated methods. $CF$ accomplished reasonable results on the pancreas segmentation, cf.~\Cref{tab:abdomenct1kcomparison}, but also delivered a surprisingly high number of self-intersecting faces on the organs in the CT test set (not so on the MRI data with identical parameters). Moreover, training $CF$ on a smaller MRI dataset did not lead to reasonable segmentation results, cf.~\Cref{tab:mrifinetuning}, and fine-tuning of a pre-trained model is prohibited by its intricate sequential training procedure. 

Eventually, we improved the robustness to shape variations and transfer learning by simplifying the architecture to its core UNet and training it with a novel deep mesh supervision technique and an additional voxel branch. 
In our experiments, \modelname is the only method that achieves a relative number of self-intersecting faces (SIF) below $0.1\%$ on all organs and both imaging modalities, cf.~\Cref{tab:abdomenct1kcomparison,tab:mrifinetuning}, while having the best segmentation Dice score on three out of four shapes on the CT data, cf.~\Cref{tab:abdomenct1kcomparison}. This indicates that \modelname attains the best trade-off between accuracy and topological correctness across different shapes and imaging modalities. The end-to-end training of \modelname further offers the opportunity to benefit from pre-training on currently emerging large corpora of annotated data; this is standard practice for conventional image segmentation but has not yet been exploited for biomedical mesh reconstruction. Its simplicity makes \modelname straightforward to implement and to use, especially in comparison to alternative approaches that combine different neural architectures (CNN + MLP for $NMF$, CNN + GNN for $V2M$/$V2C$) or stack multiple UNets ($CF$).

Another potential that previous works have overlooked is the relation between the voxel and the mesh output, e.g., obtained by $V2C$~\cite{Bongratz_2022_CVPR} and \modelname. In \Cref{tab:registration}, we demonstrated that a simple surface registration via (NR)ICP of the predicted meshes to the learned voxel segmentation improves the average (ASSD) and worst-case (HD99) surface distance by up to 2mm (spleen HD99 of \modelname with no registration and NRICP). However, in the case of NRICP, this might come at the cost of additional self-intersections, which are avoided by the rigid nature of standard ICP, still yielding a consistent improvement in ASSD for \modelname on all organs. Therefore, we recommend applying NRICP only if accuracy is more critical than the manifoldness of the shapes and ICP otherwise.

We based our implementation of all methods on the original repositories and tuned critical parameters, such as the weighting between individual loss terms, thoroughly. This was necessary since we found slight misspecification of these parameters to cause unreasonable results, and none of the evaluated methods, except $V2M$, was initially developed for abdominal shapes. However, finding the best set of parameters is time-consuming and computationally costly. Following the same procedure for all methods for a fair comparison, we limited the search to the liver shape. Nonetheless, our evaluation might be biased toward this shape.

The labels we used for training and evaluation in our experiments were obtained from manually annotated voxel masks. However, as seen from \Cref{fig:figure1}, this is not optimal due to the limited resolution of the voxels and prevalent staircase artifacts. Unfortunately, in contrast to neuroimaging, where established and extensively validated software frameworks for automated mesh extraction exist~\cite{fischl2012freesurfer}, there is no such tool for the abdomen and the creation of ``manual mesh annotations'', is hardly possible. 

In summary, we have transferred recent template-based surface extraction methods to abdominal multi-organ segmentation from CT and MRI. We have uncovered the limited generalization ability of these approaches to changes in organ geometry and a severe dependency on the amount of training data. We were able to mitigate these issues with a simpler, yet competitive deep mesh deformation architecture, \modelname. We demonstrated generalization across different abdominal organs and two imaging modalities, CT and MRI. We are encouraged by this result to question overly complex neural architectures in favor of more straightforward but similarly effective methods.

\section*{Methods}

\subsection*{Datasets}
The two imaging modalities in our experiments are CT and MRI. For CT images, we used the public AbdomenCT-1K dataset~\cite{Ma-2021-AbdomenCT-1K}, which contains 1,000 abdominal CT scans with manual annotations for four organs: kidneys, liver, spleen, and pancreas. After excluding scans with missing organs, we split the CT data into train, validation, and test splits with 666, 167, and 147 scans, respectively. CT images were z-score normalized. For MRI scans, we used Dixon opposed-phase (OPP) images from three different datasets: UKBiobank~\cite{Littlejohns2020ukb} (37 scans), GNC~\cite{nako2014} (16 scans), and KORA~\cite{Bamberg2016SubclinicalDB} (17 scans). The OPP images were pre-processed following the steps from a previous study~\cite{Rickmann2022abdomennet}, which involves bias field correction, resampling, cropping, and annotation by an anatomy expert. Our MRI training, validation, and test splits contained 47, 5, and 18 samples, respectively, balanced according to data sources. MRI images were min-max normalized. For all images, we performed affine registration~\cite{Modat2014GlobalIR} to a randomly selected reference image (Case\_00001 from AbdomenCT-1K, excluded from the test set) to ensure spatial alignment ($2\stimes2\stimes3\text{mm}^3$ resolution). Finally, we extracted ground truth meshes for each organ from the label masks using marching cubes~\cite{lewiner2003efficient}.

\subsection*{Development of a new deep diffeomorphic mesh deformation method}

\subsubsection*{Voxel space}
At the core of \modelname, we employ a 3D residual UNet based on the 3D-full-resolution network implemented in {\rev nnUNet}~\cite{isensee2020}. This fully-convolutional neural network projects a 3D input scan into a low-resolution latent space. From the latent space, the decoder recovers the original image size step-wise, producing a stack of cuboidal feature maps with increasing resolution. From the final UNet layer, a segmentation is obtained via Softmax activations.

\subsubsection*{Neural flow fields and deep mesh supervision} \label{sec:dms}
The lower part of \Cref{fig:architecture} is dedicated to the deformation of the mesh template $\mathcal{T} \subset \mathbb{R}^3$. From a high-level view, \modelname computes an (approximately) diffeomorphic deformation, i.e., an invertible mapping $f: \mathbb{R}^3 \rightarrow \mathbb{R}^3$, from the template to the output shapes. 
{\rev We keep the connectivity of the mesh fixed so that the mapping inherently creates point-wise correspondence to the template --- an important pre-requisite for statistical analyses and quantitative comparison of extracted shapes.}
In this work, we mainly deal with meshes that can be subdivided into $K$ connected components (CCs) representing $K$ individual organs. Each CC is (approximately) homeomorphic to the sphere $S^2 = \{x \in \mathbb{R}^3: \, \norm{x}_2 = 1\}$ and therefore 2-manifold. This manifoldness, however, may be broken by self-intersections.

Inspired by recent advances in neural shape deformations~\cite{gupta2020_neuralmeshflow,lebrat2021corticalflow}, we model the deformation of the mesh template via an ordinary differential equation (ODE). Formally, the trajectory of any surface point $x\in \mathcal{T}$ is described by
\begin{equation} \label{eq:ode}
    \frac{dx(t)}{dt} = \Phi(x(t), t), \quad x(t=0) = x_T,
\end{equation}
where $x_T$ is the initial location on the template, i.e., the boundary condition of the ODE. Figuratively, $\Phi$ is a flow field in Euclidean space that acts on the surface points and causes a deformation of the template. 
{\rev Importantly, there exists a unique trajectory-describing solution to \Cref{eq:ode} for each point on the template~\cite{lebrat2021corticalflow}, which implies that self-intersections are avoided in an optimal scenario. In practice, however, a numerical black-box solver must be applied to find the solution so that self-intersections might still occur due to the discretization~\cite{dupont2019_augmented}. Yet, we find this to happen rarely in our experiments, cf.~\Cref{tab:abdomenct1kcomparison}.}


While, in general, $\Phi$ is time-varying, we define it to be mostly constant and to only change at discrete equidistant time points $t_i$. Without loss of generality, we assume $t_i \in \mathbb{N}_0$, so that
\begin{equation} \label{eq:phi}
    \Phi(x(t), t) = \Phi_{\lfloor t \rfloor}(x(t)),
\end{equation}
where $\{\Phi_i:\, i = 0, \ldots, 4\}$ is a set of five stationary flow fields. In contrast to previous work~\cite{lebrat2021corticalflow}, where three UNets parameterize $\Phi_i$, we compute $\Phi_{0-4}$ directly from the feature maps of a single UNet decoder (see \Cref{fig:architecture}). This makes the network end-to-end trainable and more efficient in terms of parameters. Moreover, the ``resolution'' of each flow field is equal to the resolution of the feature map at the respective decoder stage. Thereby, the level of detail in the reconstruction is increased gradually from coarse to fine, ultimately enabling a high reconstruction accuracy and topological correctness, cf.~\Cref{tab:abdomenct1kcomparison}.

\subsubsection*{Loss function}
The overall loss function is a sum of voxel-related and mesh-related terms. From the final and deep supervision~\cite{zeng2017deepsupervision} segmentation output, i.e., $\pred{y} \in \mathbb{R}^{HWD}$, we compute the cross-entropy loss 
\begin{equation} \label{eq:cross entropy}
		\mathcal{L}_\text{CE}(\pred{y}, \gt{y}) = - \frac{1}{HWD} \sum_{i=1}^{HWD} \log \frac{\exp \pred{y}_{i, \hat{c}}}{\sum_{c=1}^{C} \exp \pred{y}_{i, c}},
\end{equation}
where $\hat{c}$ is the ground truth class of a certain voxel as specified by $\gt{y}$. In our experiments, we combine kidneys into a single label so that $C=5$ including the background.
From the final and DMS meshes, we compute the Chamfer loss
\begin{equation} \label{eq:Chamfer}
	\mathcal{L}_\text{Ch} (\Ppred, \Pgt) 
	= \frac{1}{\abs{\Pgt}} \sum_{\vec{u} \in \Pgt} \underset{\vec{v} \in \Ppred_{s}}{\min} \norm{\vec{u} - \vec{v}}^2
	+ \frac{1}{\abs{\Ppred}} \sum_{\vec{v} \in \Ppred}  \underset{\vec{u} \in \Pgt}{\min} \norm{\vec{v} - \vec{u}}^2,
\end{equation}
based on $\abs{\Pgt}=\abs{\Ppred}=50,000$ sampled surface points per organ. To improve surface regularity, we add an edge loss term
\begin{equation}
	\mathcal{L}_\text{E}(\mathcal{M}) = \frac{1}{\abs{\mathcal{E}}} \sum_{(i,j) \in \mathcal{E}} \norm{\vec{v}_i - \vec{v}_j}^2,
\end{equation}
where $\mathcal{E}$ is the set of edges of the mesh $\mathcal{M}$.
Taken together, we train \modelname with the loss function
\begin{equation}
    \mathcal{L} = \mathcal{L}_\text{CE} + \mathcal{L}_\text{Ch} + \lambda \mathcal{L}_\text{E},
\end{equation}
where each term includes the sum of final and deep supervision outputs and $\lambda = 10$.

\subsubsection*{Multi-organ segmentation} \label{sec:multi-task}
While most previous template-based surface extraction methods are limited to single shapes, \modelname can handle an arbitrary number of organs in parallel. To this end, we predict a separate deformation field for each organ in the intermediate deformation stages. In the final stage, i.e., $\Phi_4$, however, we deform all shapes with a single deformation field. As confirmed by \Cref{fig:segmentation_main}, this prevents organ boundaries from intersecting in regions where two organs lie very close together or even touch each other superficially (e.g., the liver and the right kidney in \Cref{fig:segmentation_main}). For aligning the predicted meshes to the voxel output, we explore using rigid~\cite{besl1992icp} and non-rigid~\cite{Amberg2007OptimalSN} surface registration implemented in version 3.20.1 of Trimesh~\cite{trimesh}.

\subsection*{Implementation details}
We implemented all models based on PyTorch (v1.10.0)~\cite{paszke2019pytorch}, torch-geometric (v2.0.4)~\cite{Fey/Lenssen/2019},
and PyTorch3d (v0.6.1)~\cite{ravi2020pytorch3d}. We tracked our experiments with Weights and Biases~\cite{wandb}.
We adapted all baseline methods to the task of multi-organ segmentation using the original repositories as a starting point. All models were trained with the same mesh template, which we created by overlaying all ground-truth segmentation maps from the CT training set, applying an occupancy threshold of 30\%, marching cubes~\cite{lewiner2003efficient}, and twenty steps of Laplacian smoothing~\cite{Vollmer1999_hclap}. For each method, we tuned the weighting among individual loss functions based on the liver shape with an initial log grid and a subsequent linear search and the ASSD as the target metric. For ODE-based methods, we used an Euler integration scheme with five integration steps ($h=0.2$) due to its computational efficiency.

\subsection*{Training strategy}
We trained and evaluated our models on Nvidia A100 (40GB) GPUs using mixed precision and a batch size of 4. For model training, we use the AdamW optimizer~\cite{loshchilov2018decoupled} ($\beta_1 = 0.9$, $\beta_2 = 0.999$) with a cyclic learning rate scheduler~\cite{Smith2017CyclicalLR} (base learning rate $\expnumber{1}{-4}$) and selected the best epoch based on the validation set (target metric ASSD) with a hard limit of 200 epochs (for $CF$, this limit applied to each UNet separately). We found the initialization of the output layers with zero weights crucial for training success (also for fine-tuning). All other initial weights follow PyTorch's default initialization (uniform).


\section*{Acknowledgements}
This research was conducted using the UK Biobank Resource under Application No. 34479. This research was supported by the German Research Foundation and the Federal Ministry of Education and Research in the call for Computational Life Sciences (DeepMentia, 031L0200A). The authors gratefully acknowledge the Leibniz Supercomputing Centre (www.lrz.de) for providing computational resources. Thanks also to Johannes Kaiser for creating the abdomen template.

\section*{Author contributions statement}
All authors conceived the experiments, F.B. implemented and conducted the experiments, F.B. and C.W. analyzed the results. F.B. and C.W. drafted and edited the manuscript and all authors have given final approval for the version submitted. 

\section*{Additional information}


\noindent
\textbf{Competing interests}

\noindent
The authors declare no competing interests.

The corresponding author is responsible for submitting a \href{http://www.nature.com/srep/policies/index.html#competing}{competing interests statement} on behalf of all authors of the paper. This statement must be included in the submitted article file.

\section*{Data availability}
AbdomenCT-1K data used in this work was available from the public repository (\url{https://github.com/JunMa11/AbdomenCT-1K}) after filling out the required data usage form. UK Biobank data was obtained from \url{https://www.ukbiobank.ac.uk} (application number 34479), KORA from \url{https://www.helmholtz-munich.de/en/epi/cohort/kora}, and GNC from \url{https://nako.de/}.

\bibliography{bibliography}






\end{document}